\documentclass[10pt,twocolumn,letterpaper]{article}

\usepackage[pagenumbers]{cvpr} % To force page numbers, e.g. for an arXiv version

\usepackage{graphicx}
\usepackage{amsmath}
\usepackage{amssymb}
\usepackage{booktabs}
\usepackage[ruled]{algorithm2e} % For algorithms

\usepackage{bm}
\usepackage{multirow}
\usepackage[dvipsnames]{xcolor}
\usepackage{soul}
\usepackage{subcaption}
\usepackage{wrapfig}
\usepackage{float}

\usepackage[pagebackref,breaklinks,colorlinks]{hyperref}

\newcommand{\expnumber}[2]{{#1}\mathrm{e}{#2}}

\newcommand{\R}{\mathbb{R}}

\usepackage[capitalize]{cleveref}
\crefname{section}{Sec.}{Secs.}
\Crefname{section}{Section}{Sections}
\Crefname{table}{Table}{Tables}
\crefname{table}{Tab.}{Tabs.}

\def\cvprPaperID{3834} % *** Enter the CVPR Paper ID here
\def\confName{CVPR}
\def\confYear{2023}

\begin{document}

\title{KiloNeuS: A Versatile Neural Implicit Surface Representation\\for Real-Time Rendering}

\author{ Stefano Esposito*\\
\small Hochschule Bonn-Rhein-Sieg\\
% h-brs.de\\
{\tt\small stefano.esposito@h-brs.de}
\and
 Daniele Baieri*\\
\small La Sapienza Universit\`a di Roma\\
{\tt\small baieri@di.uniroma1.it}
\and
 Stefan Zellmann\\
\small Hochschule Bonn-Rhein-Sieg\\
{\tt\small stefan.zellmann@h-brs.de}
\and
 Emanuele Rodol\`a\\
\small La Sapienza Universit\`a di Roma\\
{\tt\small rodola@di.uniroma1.it}
\and
 Andr\'e Hinkenjann\\
\small Hochschule Bonn-Rhein-Sieg\\
{\tt\small andre.hinkenjann@h-brs.de}
}
\maketitle

%%%%%%%%% ABSTRACT
\begin{abstract}
NeRF-based techniques fit wide and deep multi-layer perceptrons (MLPs) to a continuous radiance field that can be rendered from any unseen viewpoint. However, the lack of surface and normals definition and high rendering times limit their usage in typical computer graphics applications. Such limitations have recently been overcome separately, but solving them together remains an open problem. 
We present KiloNeuS, a neural representation reconstructing an implicit surface represented as a signed distance function (SDF) from multi-view images and enabling real-time rendering by partitioning the space into thousands of tiny MLPs fast to inference. As we learn the implicit surface locally using independent models, resulting in a globally coherent geometry is non-trivial and needs to be addressed during training. 
We evaluate rendering performance on a GPU-accelerated ray-caster with in-shader neural network inference, resulting in an average of 46 FPS at high resolution, proving a satisfying tradeoff between storage costs and rendering quality. In fact, our evaluation for rendering quality and surface recovery shows that KiloNeuS outperforms its single-MLP counterpart.
Finally, to exhibit the versatility of KiloNeuS, we integrate it into an interactive path-tracer taking full advantage of its surface normals. We consider our work a crucial first step toward real-time rendering of implicit neural representations under global illumination.

\end{abstract}

%%%%%%%%% BODY TEXT
\section{Introduction}
\label{sec:intro}

\begin{figure}[t]
\centering
\includegraphics[width=\columnwidth]{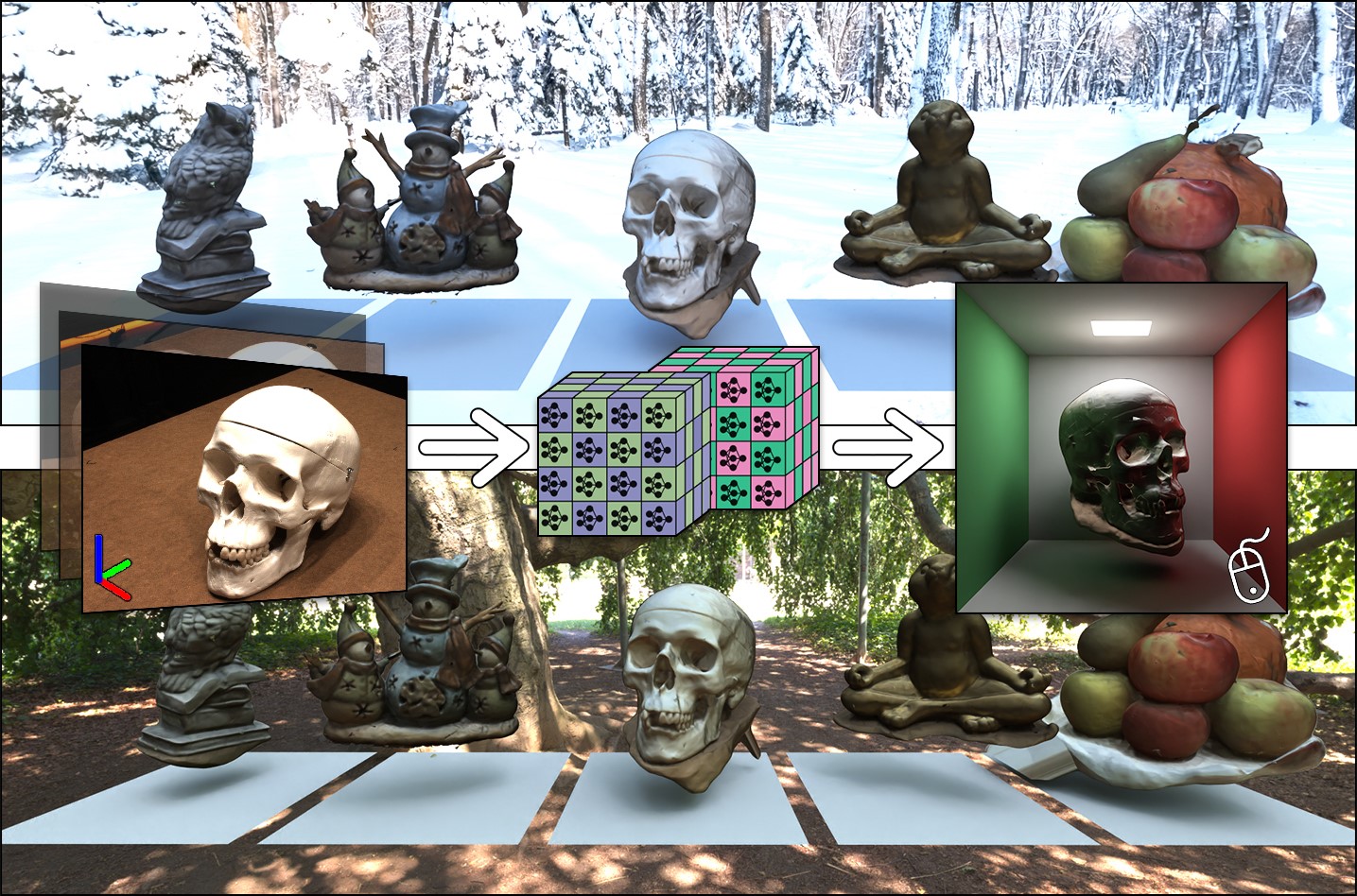}
\caption{A showcase of KiloNeuS 3D objects under different environment illumination conditions. Our neural representation continuously encodes geometry and appearance of the shapes in thousands of small neural networks.}
\label{fig:teaser}
\end{figure}

\footnotetext{* Indicates equal level of contribution.}
For a human being, the task of understanding the geometry of a 3D object shown in an image is usually easy. Unfortunately, the same cannot be said for a computer. The loss of one dimension during the projection step makes the problem ill-posed, since infinitely many different geometries could generate the same image through rendering. For this reason, estimating the true 3D underlying geometry remains a challenging problem. In recent years there has been rapid progress in inverse rendering techniques reconstructing a scene description from observations using differentiable rendering optimization. The most advanced methods in the field use neural networks to learn 3D representations as neural fields, encoding them in their weights and biases. NeRF \cite{mildenhall2020nerf} is a neural object representation that paved the way for a whole range of new techniques indirectly addressing the problem of geometry reconstruction by means of an optimization process which maximizes a similarity metric between generated scene-specific renderings and supervised viewpoints. NeRF-like techniques fit coordinate-based multi-layer perceptrons (MLPs) to a set of training images estimating volumetric fields such as view-invariant density and view-dependent color (also called \emph{radiance}). These volumes can be sampled and rendered from any virtual camera using differentiable integration of volumetric rendering algorithms, and are trained by minimizing the error between rendered and training images. %%

Computing photo-realistic interactions of light with surfaces requires having access to various surface properties such as normals, color (texture) and materials. The former, in particular, are used to bounce rays off its surface, continuing their path in the scene. As previously shown by Oechsle \etal \cite{oechsle2021unisurf}, a vanilla NeRF model defines objects as volume clouds; while this allows for extreme photo-realism in novel synthesized views, it lacks an implicit surface representation enforcing constraints on the learned density field, not guaranteeing smooth, well-defined surface reconstruction and raising a problem in this sense. Wang \etal \cite{wang2021neus} solved this problem by decoupling geometry from appearance and deriving the optical density field from a learnable signed distance function (SDF) that implicitly defines surfaces in the scene, and whose normals can be analytically computed. Representing the surfaces as signed distances rather than volume clouds also allows the optimization to converge in fewer iterations, since the function space representing the geometry is much more constrained.

Neural object representations do not provide direct access to surface geometry, which is why they are rendered by means of na\"ive ray marching-derived approaches with dense ray sampling, leading to a steep increase in rendering times.
Visualization of these representations can thus be very computationally intensive. An extremely large number of feedforward passes through a wide and deep neural network must be performed for each ray cast. Depending on the target resolution and the number of ray samples to take, a rendering of a vanilla NeRF model, for example, could take minutes to complete, which is well beyond the requirements of real-time rendering. The authors of KiloNeRF \cite{reiser2021kilonerf} demonstrated how real-time rendering can be achieved by employing thousands of small and lightweight MLPs, which uniformly partition the scene into geometrically simpler same-sized blocks, in place of a single large MLP. These can be queried at high rates, resulting in a rendering speed-up of three orders of magnitude compared to vanilla NeRF.
We believe that these rendering speedups could be increased even further by representing neural objects as SDFs, as we may resort to sphere tracing \cite{hart1993sphere} instead of volumetric ray marching. 

The combination of the capability of neural representations to capture 3D models from real-world objects in a simple, automated fashion, with the possibility to visualize them in real-time allows for a new degree of freedom in accessing and manipulating 3D data. In order to achieve this goal, we propose the following contributions:

\begin{itemize}
    \item We define KiloNeuS, a novel neural representation designed to be fast to render in standard ray-tracers, overcoming both NeRF's slow inference times -- in a similar fashion to the KiloNeRF approach -- and its lack of a well-defined implicit surface, following the NeuS scheme.
    \item We conduct a comprehensive evaluation of our model, in terms of representation quality, features and rendering efficiency, with respect to the results of the latest proposals in the field, showing that our proposal is competitive with the state of the art in this aspect.
    \item We demonstrate the potential of our neural representation by integrating it into an interactive path-tracer supporting scenes that may contain bounded objects of both classic (e.g., polygonal mesh) and neural representations. This application allows to show both the interactive performance of our method and, under some assumptions, the light interactions between objects of different types. 
\end{itemize}

\section{Related Work}

Our work intersects several lines of research, most notably neural rendering, implicit representations (and in particular, neural radiance fields), and neural scene relighting. We review works relating to ours in each of them.

\paragraph{Neural Rendering}
Inverse rendering refers to a set of techniques estimating intrinsic scene characteristics such as camera, geometry, material and light parameters, given a single image or a set of images using differentiable rendering optimization \cite{marschner1998inverse, deschaintre2019flexible, henzler2019escaping, deschaintre2018single, li2018materials}.
Neural rendering is closely related to inverse rendering; it introduces components learned by deep neural networks into the rendering pipeline in place of predefined physical models or data structures used in classical rendering. A typical neural rendering approach takes as input scene properties and builds a neural scene representation from them, which can then be rendered with different properties to synthesize novel images. The authors of NeuS \cite{wang2021neus} proposed to decouple geometry from appearance by jointly learning a) a signed distance function (SDF), which analytically defines the density field and the surface normals, and b) a color field defining the surface appearance.

\begin{figure}[h]
\begin{minipage}[b]{0.45\linewidth}
\centering
\includegraphics[width=1.0\linewidth]{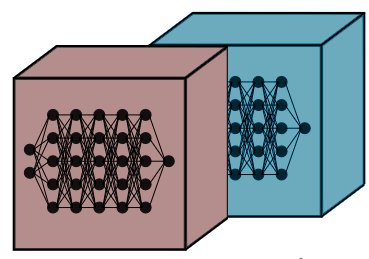}
\end{minipage}
\hspace{0.5cm}
\begin{minipage}[b]{0.45\linewidth}
\centering
\includegraphics[width=1.0\linewidth]{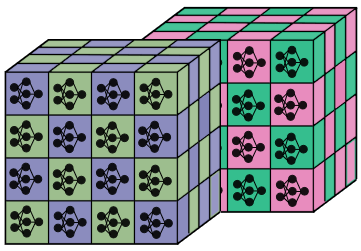}
\end{minipage}
\caption{High-level visualization of the structural difference between the NeuS model (left), composed of two large MLPs, and our KiloSDF and KiloColor grids of small MLPs (right).
}
\label{fig:kilonerf_pic}
\end{figure}

\paragraph{Implicit Representations}
Representing 3D surfaces as continuous functions over space is a recent trend in geometric deep learning. Manipulating continuous surfaces offers new perspectives in geometry processing, and led to significant advances in various complex tasks. Occupancy fields \cite{Occupancy_Networks} and signed distance fields \cite{Park_2019_CVPR} found major applications in 3D shape reconstruction \cite{Peng2020ECCV, Atzmon_2020_CVPR, icml2020_2086, sitzmann2019siren} and 4D shape reconstruction \cite{Lei2022CaDeX, OccupancyFlow, chen2021snarf, tiwari21neuralgif}. Moreover, several recent proposals tackle 3D surface recovery from multi-view images using SDFs \cite{oechsle2021unisurf, yariv2020multiview, DVR}. 
In addition, signed distance functions allow to render 3D scenes by sphere tracing \cite{hart1993sphere}, a ray-marching heuristic allowing for real-time rendering. 

\paragraph{Neural Radiance Fields}
Neural radiance fields were proposed in \cite{mildenhall2020nerf} and gained much attention from the community because of their unprecedented results on novel view synthesis of 3D scenes, the problem of generating novel camera perspectives of a scene given a fixed set of images. Follow-up works attempted to improve and extend the original proposal: \cite{wang2021nerfmm, meng2021gnerf} proposed unsupervised (i.e. without camera parameters) formulations for NeRF training, while \cite{dnerf, tretschk2020nonrigid, park2021nerfies} generalized the setting to non-rigidly deforming scenes.

\paragraph{Speeding-up Neural Rendering}
Several approaches have been proposed to speed up the rendering of NeRF-based representations. 
NSVF \cite{liu2020neural} progressively learns an underlying structure of voxel-bounded implicit fields organized in a sparse voxel octree to model local properties in each cell.
Using a different approach, the authors of AutoInt \cite{lindell2021autoint} proposed a technique learning closed-form solutions to integrals by fitting a network representing the antiderivative. While this can be used in a variety of contexts, the authors claim it can speed up NeRF numerical approximation of the volume rendering integral of an order of magnitude by significantly reducing the number of queries needed per pixel.
DONeRF \cite{neff2021donerf}, on the other hand, jointly trains a depth oracle network to perform only sample points placed locally around surfaces in the scene, reducing the inference costs by up to $\sim 48$x compared to NeRF reaching interactive framerates ($\sim 20$ FPS) and visual quality. 
On a different note, DeRF \cite{rebain2021derf} proposed a spatial decomposition of a scene based on Voronoi cells, where each part is rendered independently and the final image is composed via Painter’s Algorithm, providing up to $\sim 3$x more efficient inference than NeRF.
Other approaches like FastNeRF \cite{fastnerf} and PlenOctree \cite{yu2021plenoctrees} speed up the volume rendering process by making use of smart \emph{caching} techniques to entirely skip the MLP query and reach impressive framerates ($> 200$ FPS), at the cost of losing the continuous representation and of a larger memory footprint that grows cubically with the targeted quality.
On a different note, Light Field Networks \cite{sitzmann2021lfns} represent a 3D scene with a 360-degree, four-dimensional light field. Such representation can render a ray with a single MLP evaluation; however, the lack of dense depth information makes it difficult to include in a path-traced scene.

\begin{figure}[h]
\centering
%\begin{subfigure}{\textwidth}
\includegraphics[width=\linewidth]{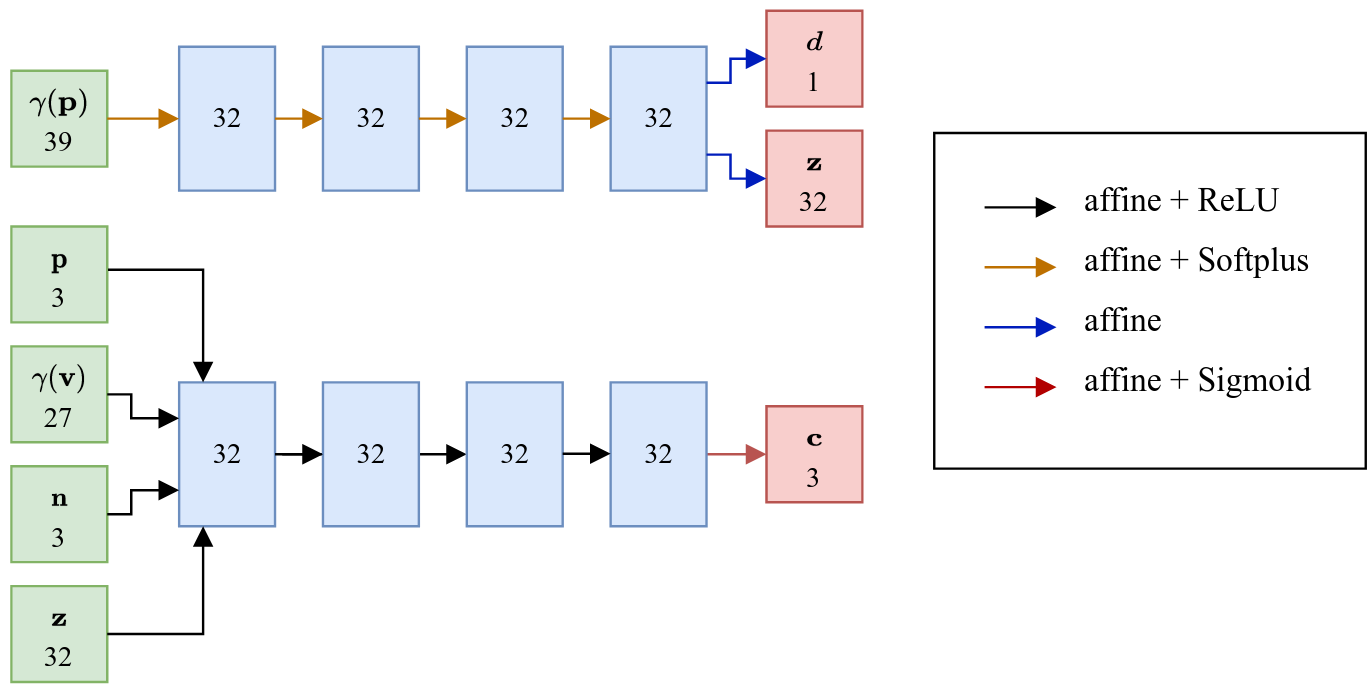}
\caption{Single KiloSDF and KiloColor MLPs architectures, designed to be a significant simplification of the respective NeuS models.}
\label{fig:kiloneus_mlp}
%\end{subfigure}
\end{figure}

Lastly, the authors of KiloNeRF \cite{reiser2021kilonerf} improved rendering times of vanilla NeRF by proposing to partition space by a uniform coarse grid of small MLPs which can be queried at higher rates than NeRF's large MLP, achieving real-time performance without caching. On a similar note, Müller \etal obtained great resonance with their Instant Neural Graphics Primitives work \cite{mueller2022instant}, mainly due to their results in extremely fast training convergence. However, they also achieve fast rendering by using a single small MLP taking as input their proposed spatial hash encoding.

%------------------------------------------------------------------------
\section{Method}

We now discuss how we achieve real-time rendering of neural object representations, outlining our contributions on top of NeuS \cite{wang2021neus} and KiloNeRF \cite{reiser2021kilonerf}. We first introduce our lightweight neural object representation, encoding implicit surfaces as signed distance fields with high fidelity, then we describe our path-tracing application supporting the aforementioned representation, along with classic computer graphic primitives in shared scenes, all in real-time.

\subsection{KiloNeuS}
\label{sec:kiloneus}

This section illustrates the KiloNeuS neural object representation, which combines the reconstructive capabilities of NeuS \cite{wang2021neus} and KiloNeRF's ability to render complex models in real-time \cite{reiser2021kilonerf}. 
KiloNeuS assumes a fixed bounding box for the learned scene and subdivides the corresponding region of space in a regular 3D grid, where each voxel is assigned two small MLPs, one learning the SDF and one the color (as shown in Figure~\ref{fig:kilonerf_pic}). In the following we refer to the aggregate SDF and color models as KiloSDF and KiloColor, which respectively replace the large and wide SDF and color MLPs used in NeuS.

Each of our MLPs only represents a smaller -- and therefore simpler -- fraction of the respective field, which requires fewer parameters to represent, thus drastically reducing inference times. Similarly to how KiloNeRF’s MLP architecture is a downscaled version of NeRF’s architecture, KiloSDF and KiloColor (Figure~\ref{fig:kiloneus_mlp}) are designed to be a downscaled version of the respective NeuS model architectures.
We also apply \emph{Fourier feature mapping} \cite{fourier_mapping} on our networks inputs, as it was shown multiple times \cite{sitzmann2019siren, mildenhall2020nerf} to sensibly improve the ability to represent high-frequency signals in coordinate-based neural networks. This positional encoding is defined as:

\begin{equation}
    \gamma_L(x) = (\hat{\gamma}_L(x_0), \hat{\gamma}_L(x_1), \dots, \hat{\gamma}_L(x_n)),
\end{equation} 
where:
\begin{multline}
    \hat{\gamma}_L(p) = \left(\sin \left(2^{0} \pi p\right), \cos \left(2^{0} \pi p\right), \dots, \right.\\ 
     \left. \sin \left(2^{\mathrm{L}-1} \pi p\right), \cos \left(2^{\mathrm{L}-1} \pi p\right)\right)
    \label{eq:freq_encoding}
\end{multline}
So that each $\R^n$ point will be encoded as a $\R^{n + 2nL}$ vector. Our KiloSDF MLPs  take as only input the 6-frequencies ($L=6$) positional encoding of spatial coordinates $x$, and output the SDF values $d$ and a feature vector $z$. On the other hand, the KiloColor MLPs require non-encoded $x$, viewing direction (encoded with $L=4$), SDF normals $n={\nabla d}/{\lVert\nabla d\rVert}$, and $z$. Following \cite{wang2021neus}, we use Softplus as activation function for KiloSDF MLPs, ensuring we will learn a smooth surface, and ReLU for KiloColor MLPs, for easier optimization and additional freedom. The outputs of KiloColor are RGB tuples, therefore they are normalized in $[0;1]$ with a Sigmoid application. 

Finally, we require a map selecting which MLP in the grid will compute output functions for a given point $x$. We identify this as $m\colon\R^3\to [N]^3$, where $N$ is the MLP grid resolution.
In the following, we identify our SDF and Color models as
\begin{equation}
    d(x) = f\left(\gamma_6(x); \bm{\Theta}_{m(x)}^d\right)
\end{equation}
\begin{equation}
    c(x, v, n, z) = \sigma\left(g\left(x, \gamma_4(v), n, z; \bm{\Theta}_{m(x)}^c\right)\right)
\end{equation}
Where $f, g$ are parametric functions for the SDF and Color MLPs and $\bm{\Theta}^{\{c, d\}}_i$ represents the learnable parameters.

Our scene model also includes a NeRF++ network \cite{kaizhang2020nerfpp} and a standard deviation trainable parameter $s$ (as in \cite{wang2021neus}). The former has been observed to be extremely useful at training time when mask supervision is unavailable: by learning the background of the scene, it helps the optimization process to disentangle the foreground object from its surrounding environment.
The latter, on the other hand, controls the standard deviation in the S-density function $\phi_s(f(\mathbf{x}))$ \cite{wang2021neus} used to train the SDF network, with ${1}/{s}$ approaching zero as the network training converges. While they are essential during training, it is important to point out that both the background model and the deviation parameter can be completely ignored when rendering the foreground object on its own with a surface rendering algorithm, since it is completely encoded in the KiloSDF and KiloColor nets.

\begin{figure}[h]
    \centering
    \includegraphics[width=.48\columnwidth]{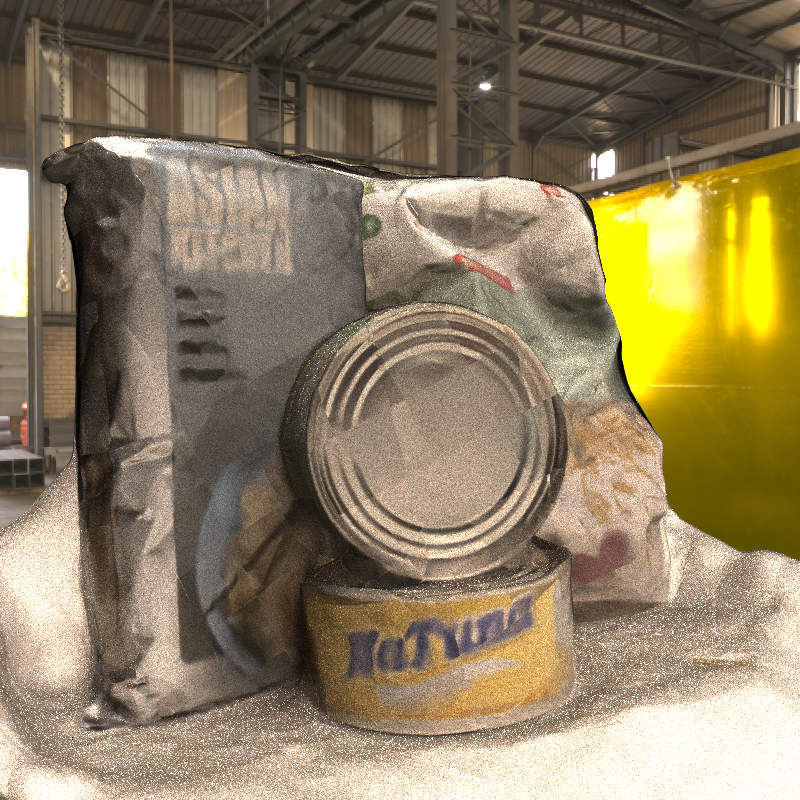}
    \hfill
    \includegraphics[width=.48\columnwidth]{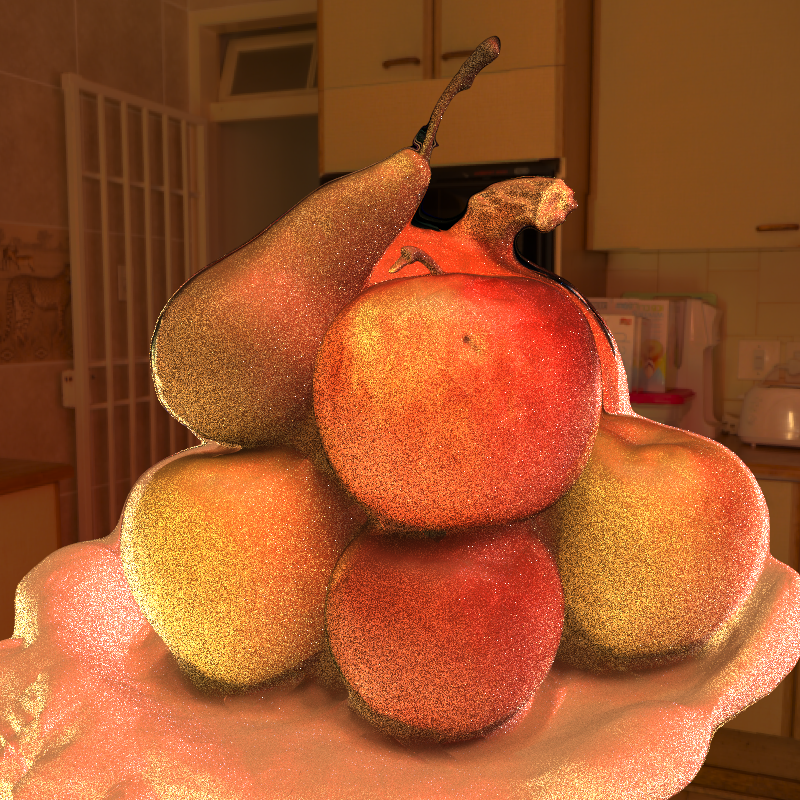}
    \caption{Path-traced renders of KiloNeuS objects (assumed to have Lambertian materials). 30 samples per pixel, no denoising pass.}
    \label{fig:kiloneus_pt_renders}
\end{figure}

\subsection{Training}
We follow the training procedure of \cite{reiser2021kilonerf} for our model optimization: first, we learn a good weight initialization through a phase of distillation training, then we refine our solution through fine-tuning. Reiser \etal \cite{reiser2021kilonerf} already observed that Kilo-models fail to converge without a distillation phase, and we found that the independence of the single MLPs, coupled with the requirement to globally reconstruct a well-defined surface, actually exacerbates this issue. We describe each phase individually.

\subsubsection{Distillation}\label{sec:distill}

In this phase, we assume to have a pre-trained NeuS model $(\hat{d}, \hat{c}, b, \nu)$ as a teacher. We optimize our model using a simple distillation loss
\begin{equation}
    L_{\text{distill}} =    \lVert \hat{d}({x}) - d({x}) \rVert_1 + \left\lVert \hat{c}({x}, {v}, \hat{n}, \hat{z}) - c({x}, {v}, \hat{n}, {z}) \right\rVert_2^2
\end{equation}
Where $x\in\R^{m\times 3}$ are sampled uniformly from the Kilo-grid bounding box, $\hat{n}\in\R^{m\times 3}$ are normalized gradients of $\hat{d}$ evaluated at $x$, and $v\in\R^{m\times 3}$ are sampled uniformly from the {visibility hemisphere} at each SDF normal in $\hat{n}$. Observe that the teacher and student color networks use different feature vectors, as these are output by the respective SDF networks and vary in dimensionality (thus, the feature vectors are not explicitly trained in this phase).

Our model is trained for 100k steps (about 3 hours in our experimental setting) using the Adam \cite{adam} optimizer, with an initial learning rate of $\expnumber{1}{-3}$ and a step learning rate schedule, with a step size of 10k and a $\gamma$ factor of $0.75$. The Kilo-grid bounding box is set to $[-1; 1]^3$.

\begin{figure*}[h]
    \centering
    \includegraphics[width=\textwidth]{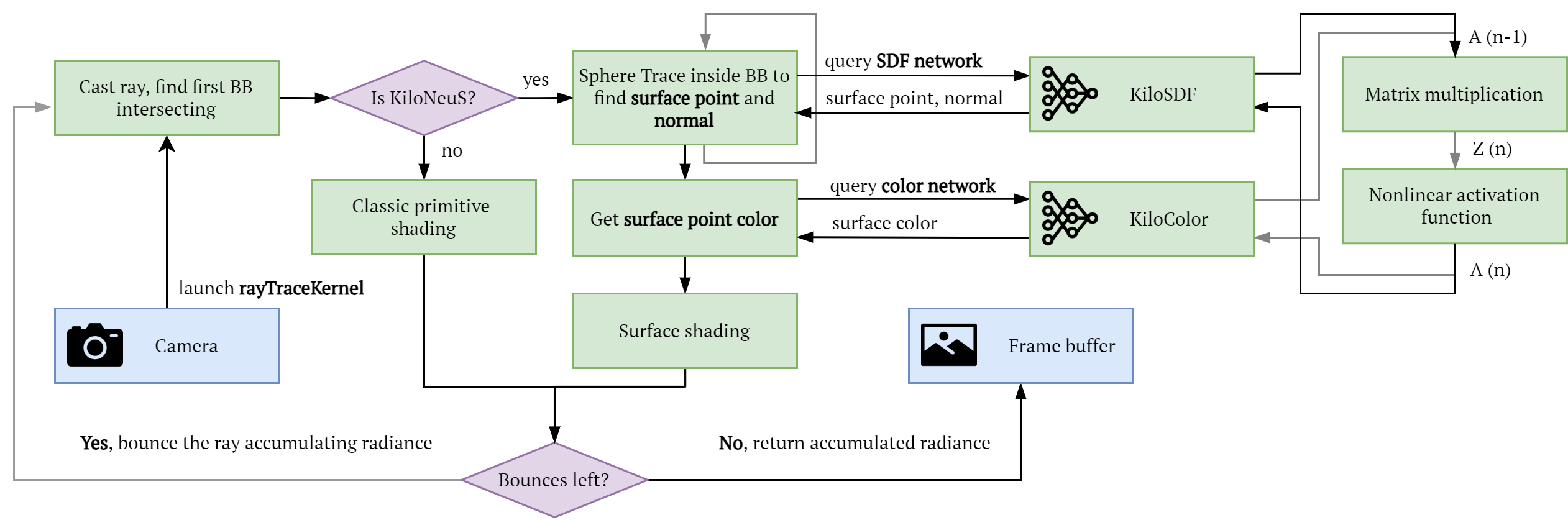}
    \caption{High-level diagram of the cuNPT GPU thread execution flow focusing on the rendering of a KiloNeuS object.}
    \label{fig:cuNPT_flow}
\end{figure*}

\subsubsection{Fine-tuning}\label{sec:finetune}

In the fine-tuning phase, we train KiloNeuS using the standard photometric loss used for radiance fields, along with additional regularizers. 
\begin{wrapfigure}{r}{0.45\columnwidth}
\hfill
  \begin{center}
    \includegraphics[width=0.45\columnwidth]{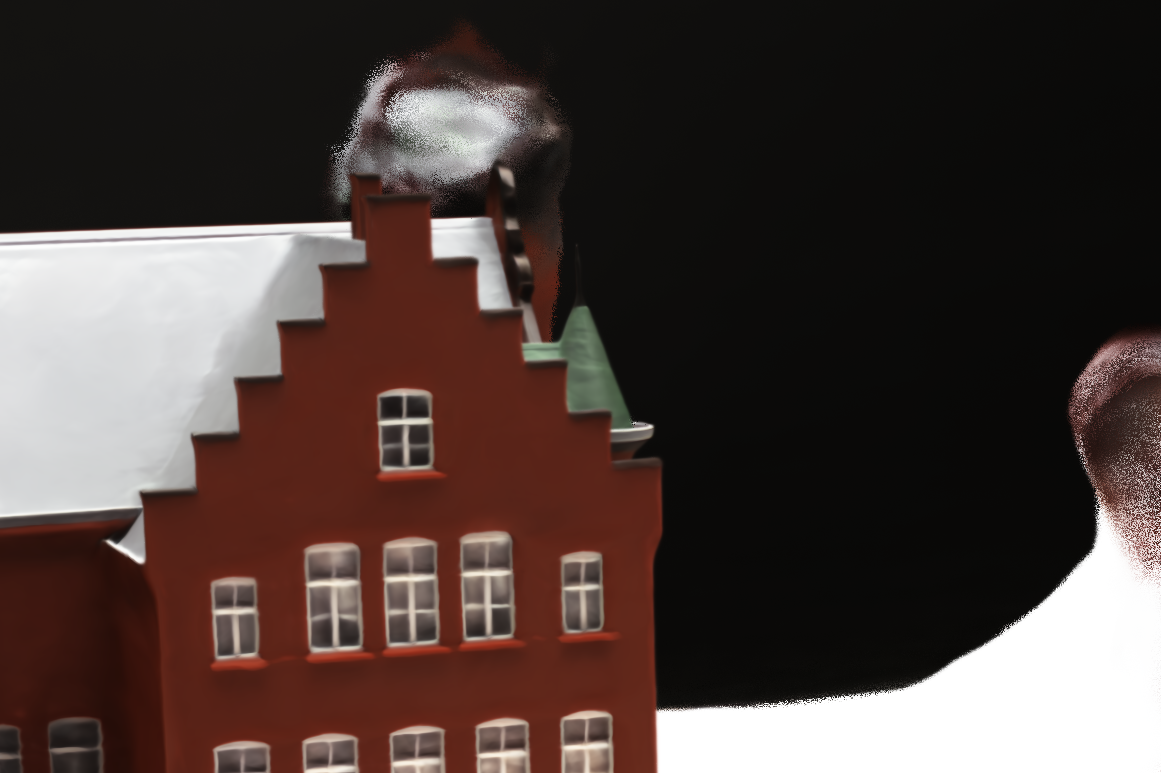}
    \includegraphics[width=0.45\columnwidth]{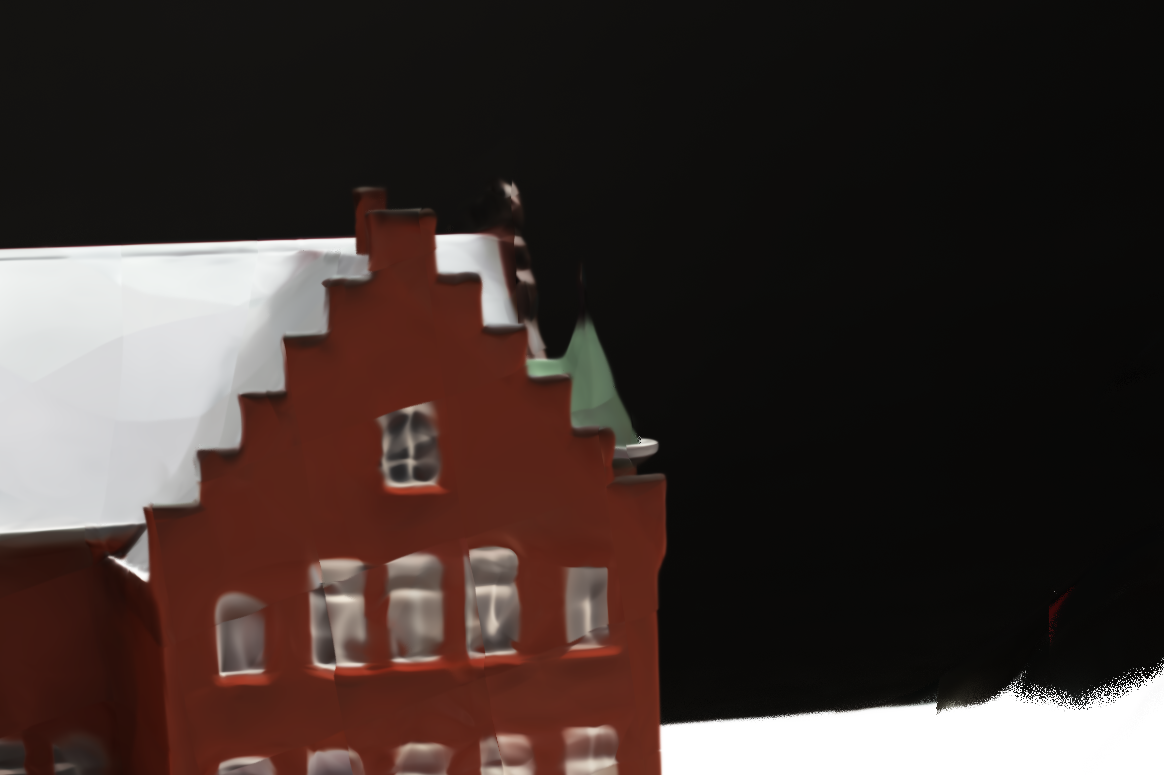}
    \includegraphics[width=0.45\columnwidth]{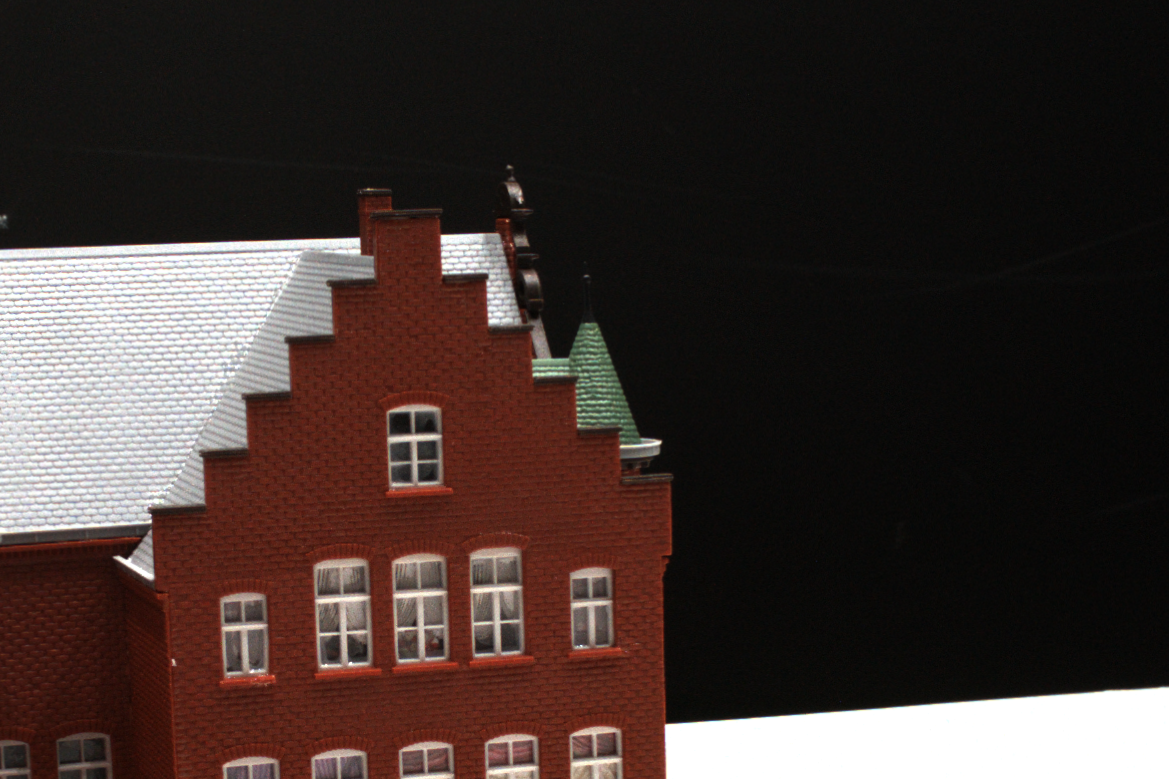}
  \end{center}
  \caption{Rendering artifacts in the NeuS representation (top), KiloNeuS improvement (center), and ground-truth image (bottom).}
  \label{fig:neus_artifacts}
\end{wrapfigure}
We use the pre-trained background model from the NeuS teacher and allow it to fine-tune as well, while the deviation network is re-initialized. 
The renderer is adapted from Wang \etal \cite{wang2021neus}, which converts SDF values to densities in order to perform differentiable volumetric rendering. We modify their solution to  allow our renderer to query the model outside the unit sphere: while this is a useful regularization when learning a scene from scratch, our distillation phase makes it unnecessary here. Furthermore, we found that NeuS occasionally creates visible geometric artifacts outside the unit sphere, which our fine-tuning phase is actually capable of removing (see Figure~\ref{fig:neus_artifacts}).

KiloNeuS is fine-tuned by considering one random camera-pose-labeled image $(I^t, (\theta_t, \eta_t))$ at a time, sampling $B$ pixels $P = \{(i_l, j_l)\}_{l=1}^B$, and applying the photometric loss:
\begin{equation}
    L_{\text{color}} = \dfrac{1}{B}\sum_{(i,j)\in P}\left\lVert C\left(d, c, o_{i, j}^t
    \right) - I^t_{i, j} \right\rVert_1
\end{equation}
where $o_{i, j}^t$ are the origin positions for tracing rays, determined in world space by $(i, j)$ and $(\theta_t, \eta_t)$, and $C$ abstracts the renderer as a function.

In order to learn a well-defined surface, one has to ensure that $d$ is actually a SDF. Following Wang \etal, we employ the eikonal loss (originally proposed by Gropp \etal \cite{icml2020_2086}):
\begin{equation}
    L_{\text{eikonal}} = \dfrac{1}{M}\sum_{i=1}^{M}\left( \left\lVert\nabla_{x_i} d(x_i)\right\rVert_2  - 1 \right)^2 
\end{equation}
We leave additional remarks regarding this regularizer for Section~\ref{sec:impl}. Finally, the complete loss function is:
\begin{equation}
    L = L_{\text{color}} + \lambda L_{\text{eikonal}} 
\end{equation}
Our model is trained in the fine-tuning phase for 100k additional steps (requiring around 6 hours in our experimental setting), using the Adam \cite{adam} optimizer, with a batch size of $B=1024$, an initial learning rate of $\expnumber{1}{-4}$ and a cosine decay schedule with $\alpha=0.05$, \ie, the learning rate would converge to $\expnumber{5}{-6}$ at the last iteration. To compute $L_{\text{eikonal}}$, we sample $M=16384$ points uniformly at random from the Kilo-grid bounding box (which we set to $[-1; 1]^3$) and we set $\lambda = 0.01$.

\begin{figure}[H]
    \centering
    \includegraphics[width=.32\columnwidth]{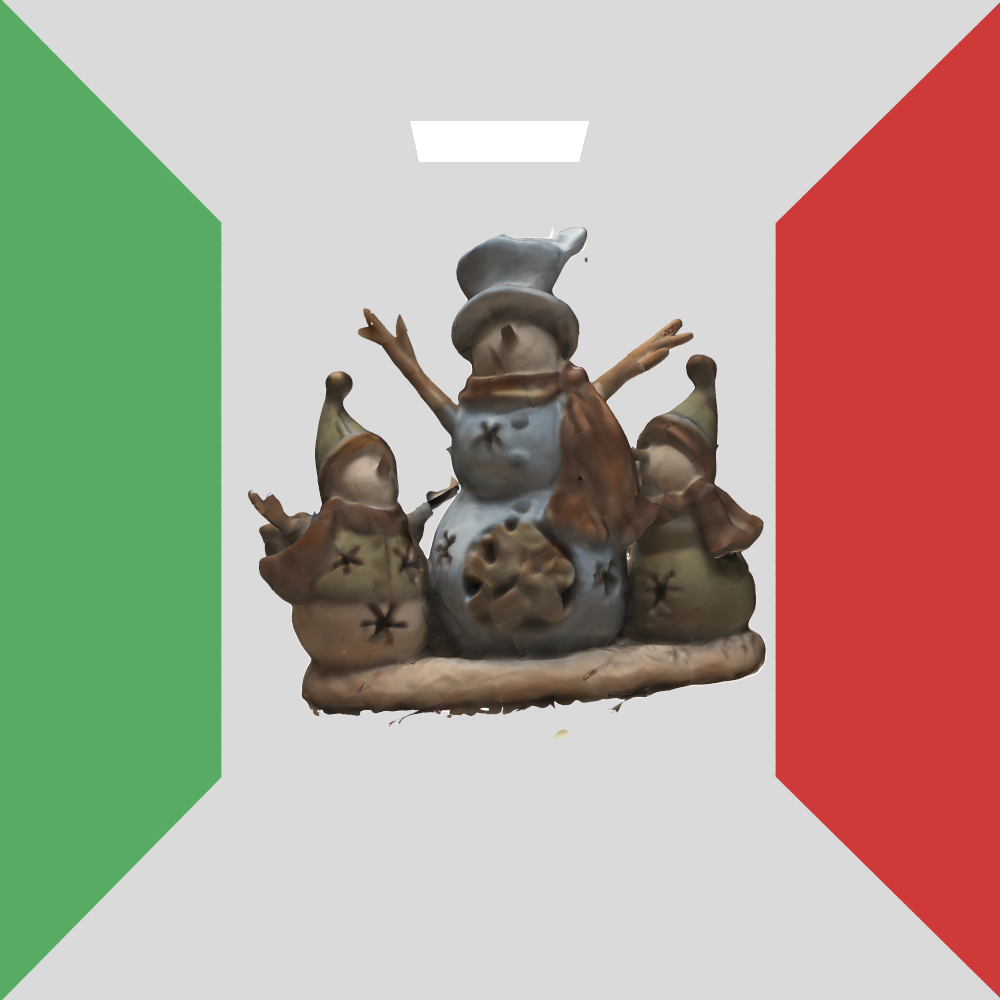}
    \hfill
    \includegraphics[width=.32\columnwidth]{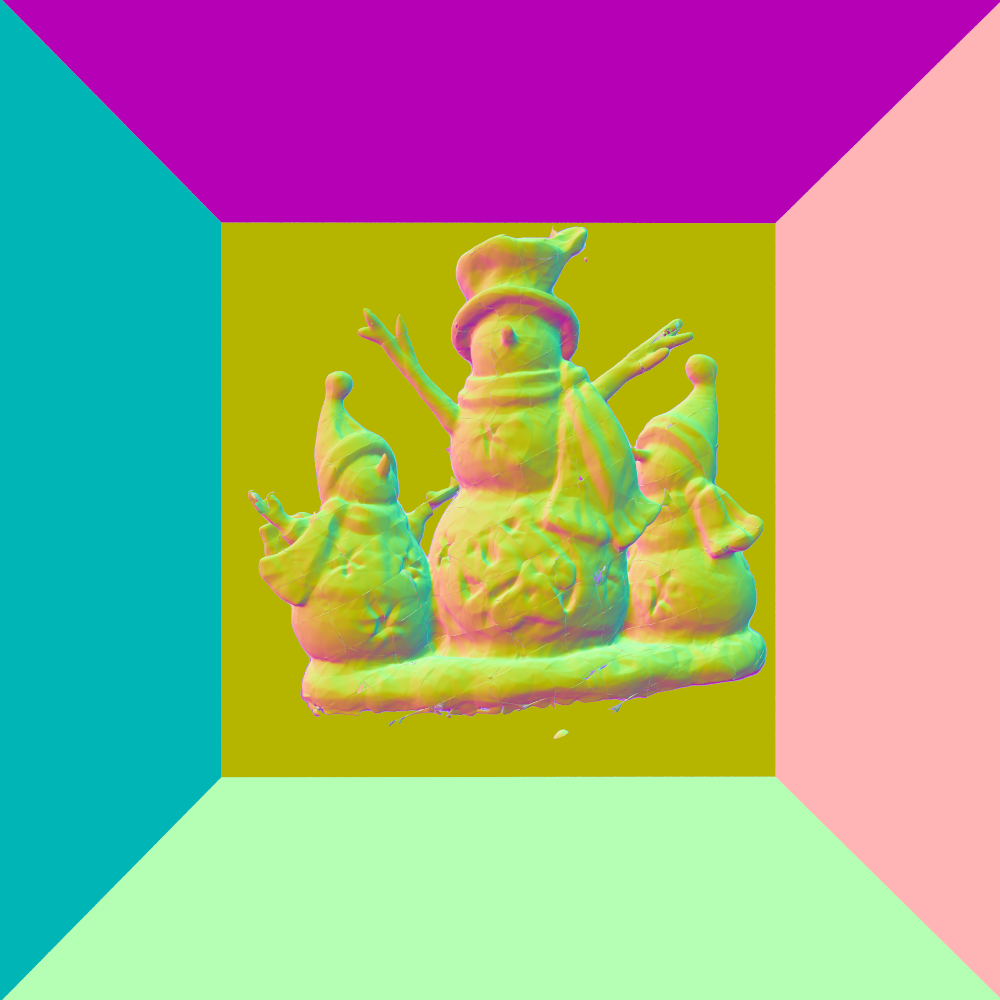}
    \hfill
    \includegraphics[width=.32\columnwidth]{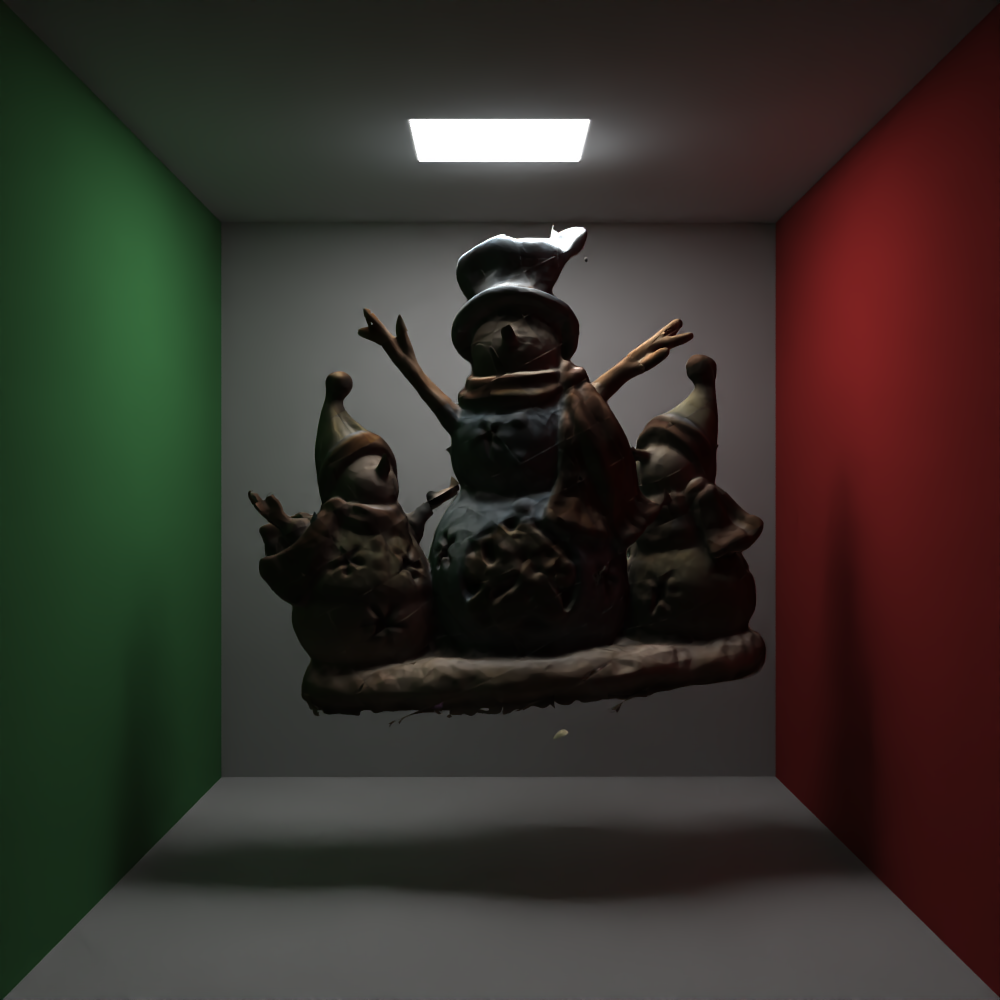}
    
    \smallskip
    
    \includegraphics[width=.32\columnwidth]{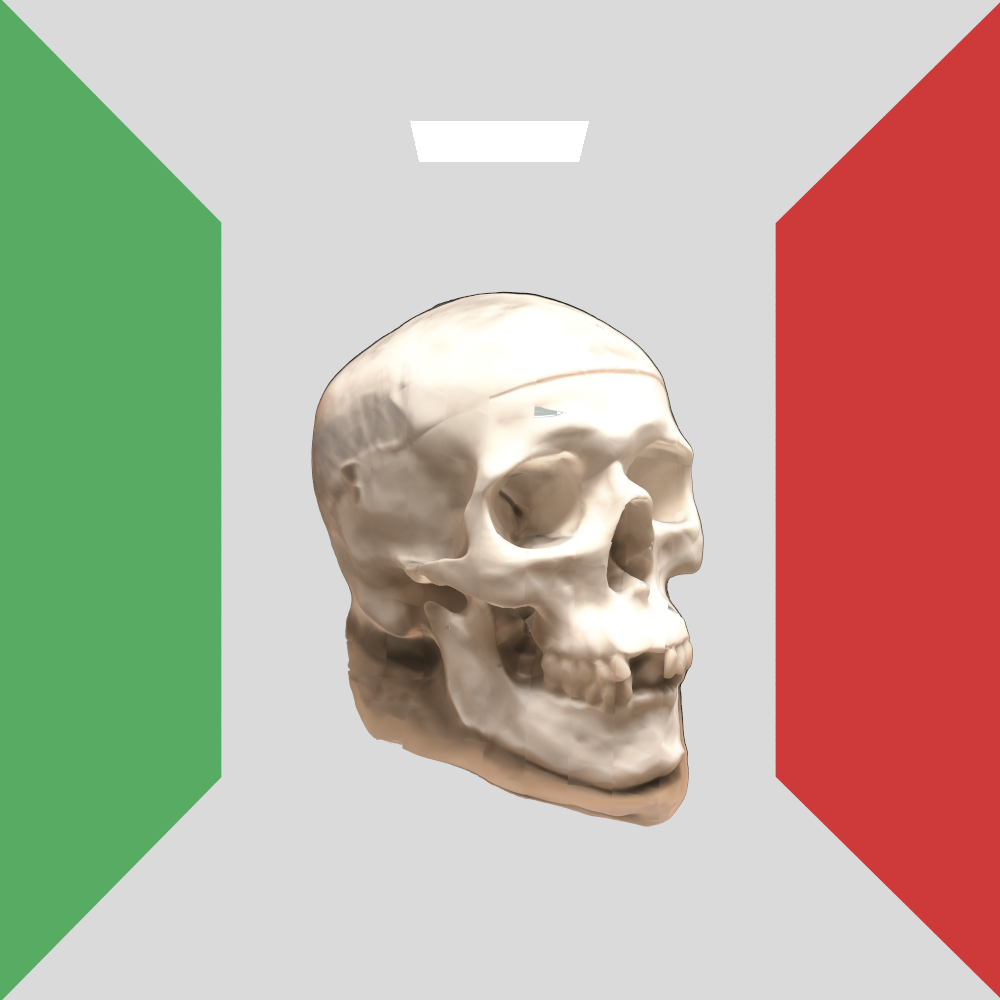}
    \hfill
    \includegraphics[width=.32\columnwidth]{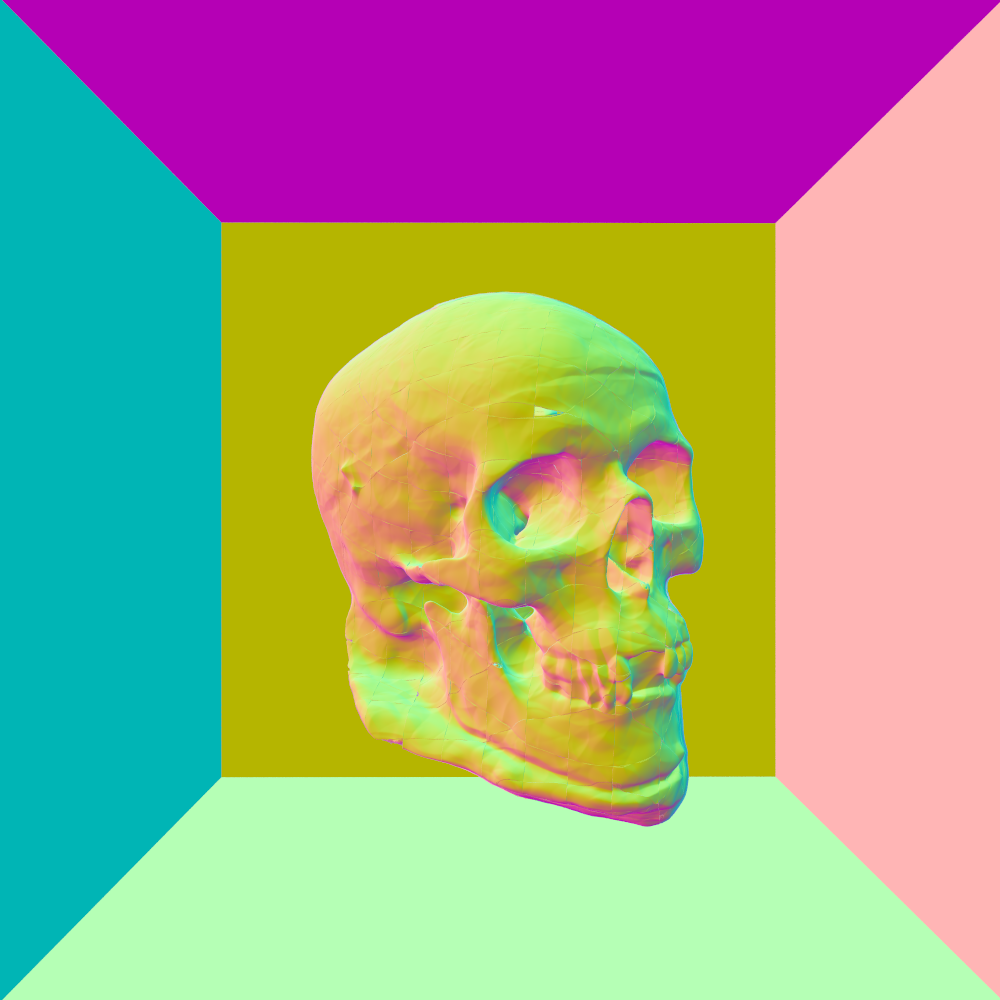}
    \hfill
    \includegraphics[width=.32\columnwidth]{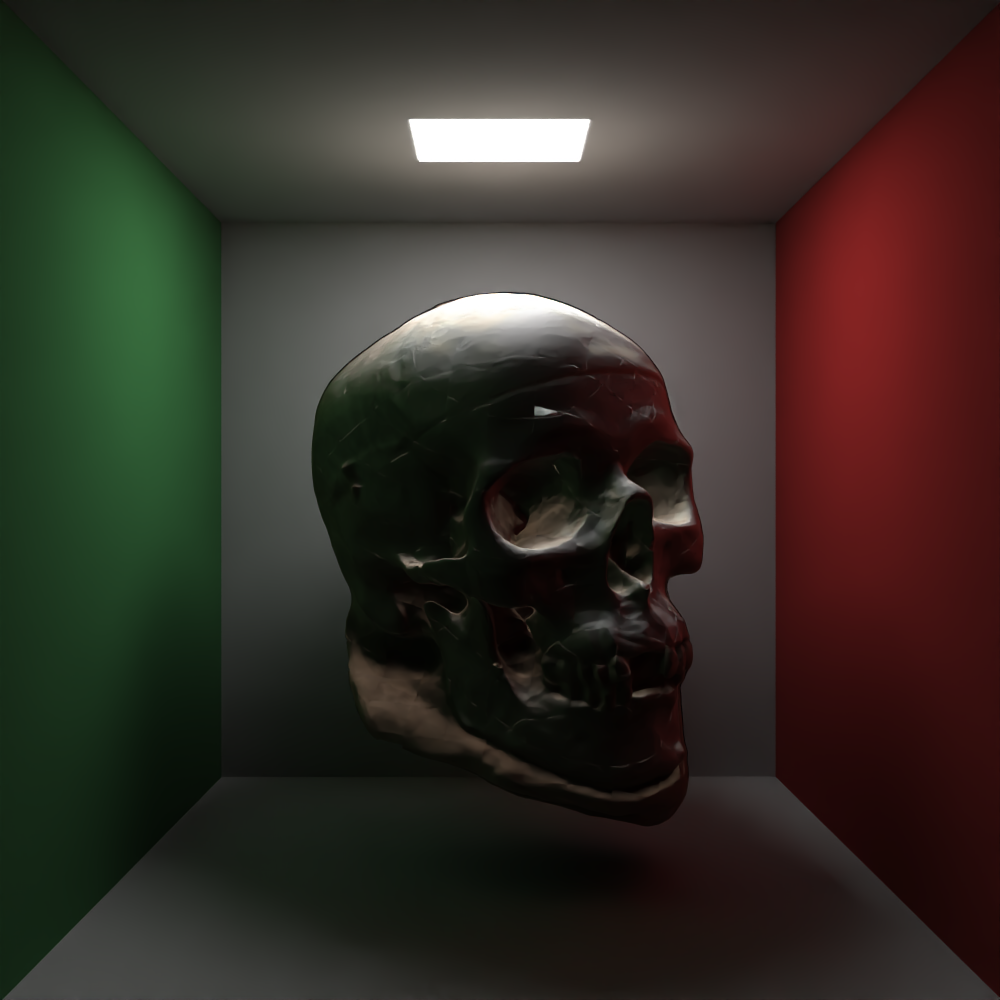}
    \caption{Left to right: color map, normal map, and path-traced render of KiloNeuS models in the Cornell box scene.}
    \label{fig:cornellbox}
\end{figure}

\subsection{Implementation details}\label{sec:impl}

In contrast to previous methods optimizing for differential properties of neural networks \cite{icml2020_2086, atzmon2021sald, wang2021neus} which achieve this goal by recursive backpropagation passes through network gradients computations, we find that finite differences is a better choice for our setting. Given that our SDF/color functions are piece-wise defined by independent MLPs, computing gradients with \textit{autodiff} would result in discontinuous gradients: this would lead to discontinuity artifacts in the output, and differential properties would be optimized for the single MLP functions rather than the global SDF/color function. We find that, in our case, optimizing $L_{\text{eikonal}}$ with finite differences allows us to control geometric discontinuity artifacts learned during the distillation phase far better than the \textit{autodiff} version (see Figure~\ref{fig:seams_artifact}).

As in Reiser \etal \cite{reiser2021kilonerf}, we develop a parallel MLP querying procedure in order to speed-up rendering at training and evaluation time. In contrast with their solution, which required the separate development of a CUDA extension, we manage to implement the procedure in Python using the \texttt{vmap} function of \textit{functorch} \cite{functorch2021}. This feature allows to automatically vectorize user-defined point-wise operations: thus, we use it to vectorize the operation of querying a particular $i$-th MLP for a batch of $m$ points, with \texttt{vmap} taking care of performing parallel invocation. The resulting function evaluates $n$ selected MLPs on $n$ batches of $m$ points.

\subsection{cuNPT: CUDA Neural Path-Tracer}\label{sec:cunpt}

Our real-time path-tracer supporting the previously discussed KiloNeuS neural object representation as well as other classic computer graphics primitives has been built upon a publicly available and parallel CUDA implementation \cite{allen_2021} of \cite{shirley}. In the following, we describe the extensions made to the rendering kernel to support KiloNeuS models, for which a high-level diagram is shown in Figure~\ref{fig:cuNPT_flow}.

\subsubsection{Rendering kernel}

For each ray, the rendering kernel finds the closest intersection point with an object bounding volume, if it exists. If the bounding box does not contain a KiloNeuS object, it follows the execution flow described in \cite{shirley}; otherwise, it starts sphere-tracing in the bounding volume intersection interval, looking for the first ray intersection with the implicitly defined neural surface. If sphere-tracing converges on the SDF 0-level set, the rendering kernel then computes its normal as the normalized SDF gradient and its color value by querying the color network at the point of intersection.
The KiloNeuS neural object representation employed in this case study does not include material properties, which at this time the path-tracer assumes to be Lambertian.

\subsubsection{Neural Network inference}

The neural network inference step is performed by CUDA device functions that take as input the spatial coordinates of a point, encode it, and perform a series of matrix multiplications and apply non-linear activation functions (whose results are always one-dimensional vectors) to output either a SDF value or a color value. Intermediate results are stored in a per-thread register memory of constant size, since global memory access would lead to extreme performance degradation. This memory, depending on the GPU used, usually has a limited capacity. The use of small MLPs with small intermediate results also helps in this regard.

\begin{figure}[h]
\captionsetup[subfigure]{labelformat=empty}
     \begin{subfigure}[b]{0.15\textwidth}
         \centering
         \includegraphics[width=\textwidth]{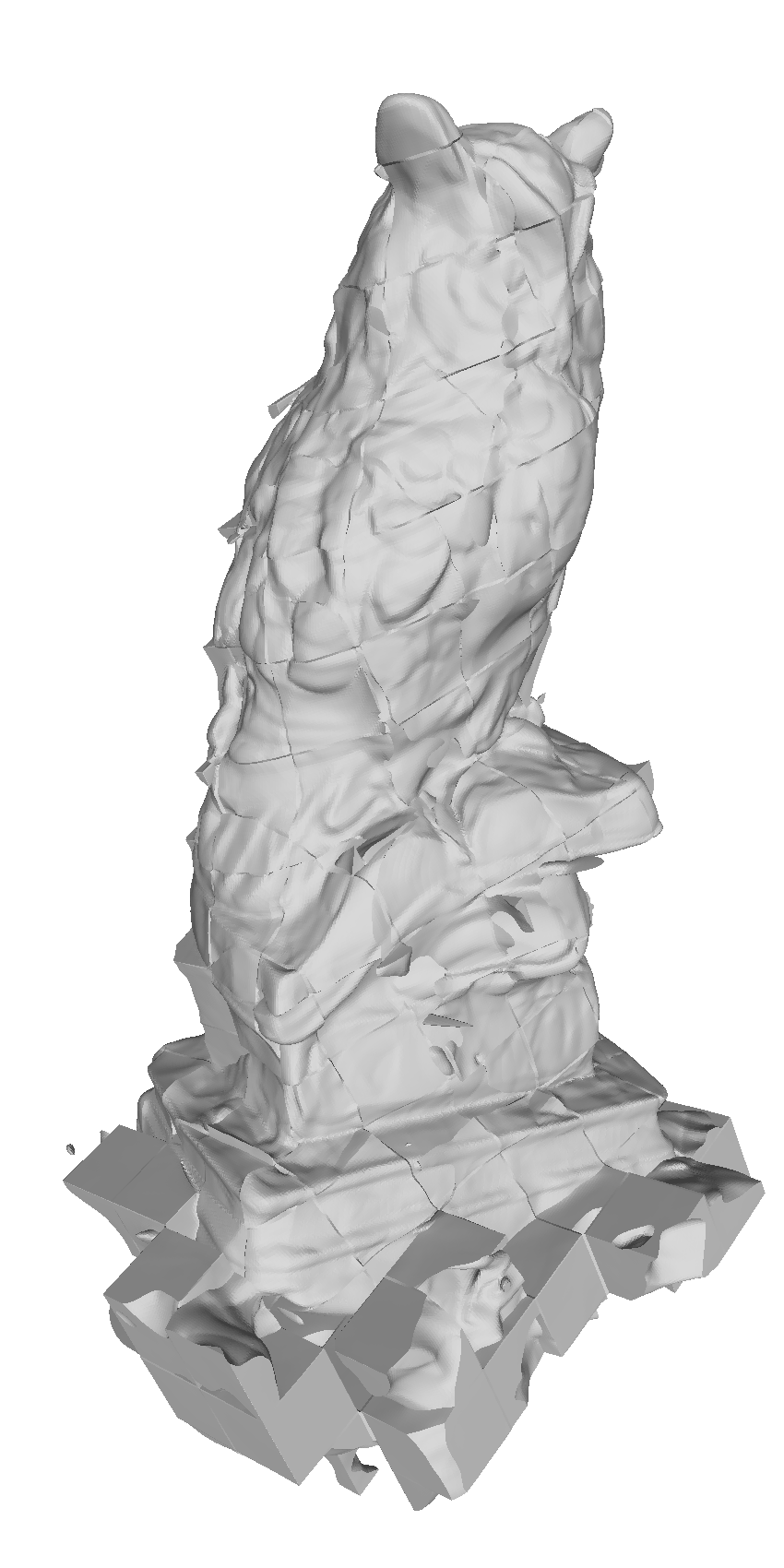}
     \end{subfigure}
     \hfill
     \begin{subfigure}[b]{0.15\textwidth}
         \centering
         \includegraphics[width=\textwidth]{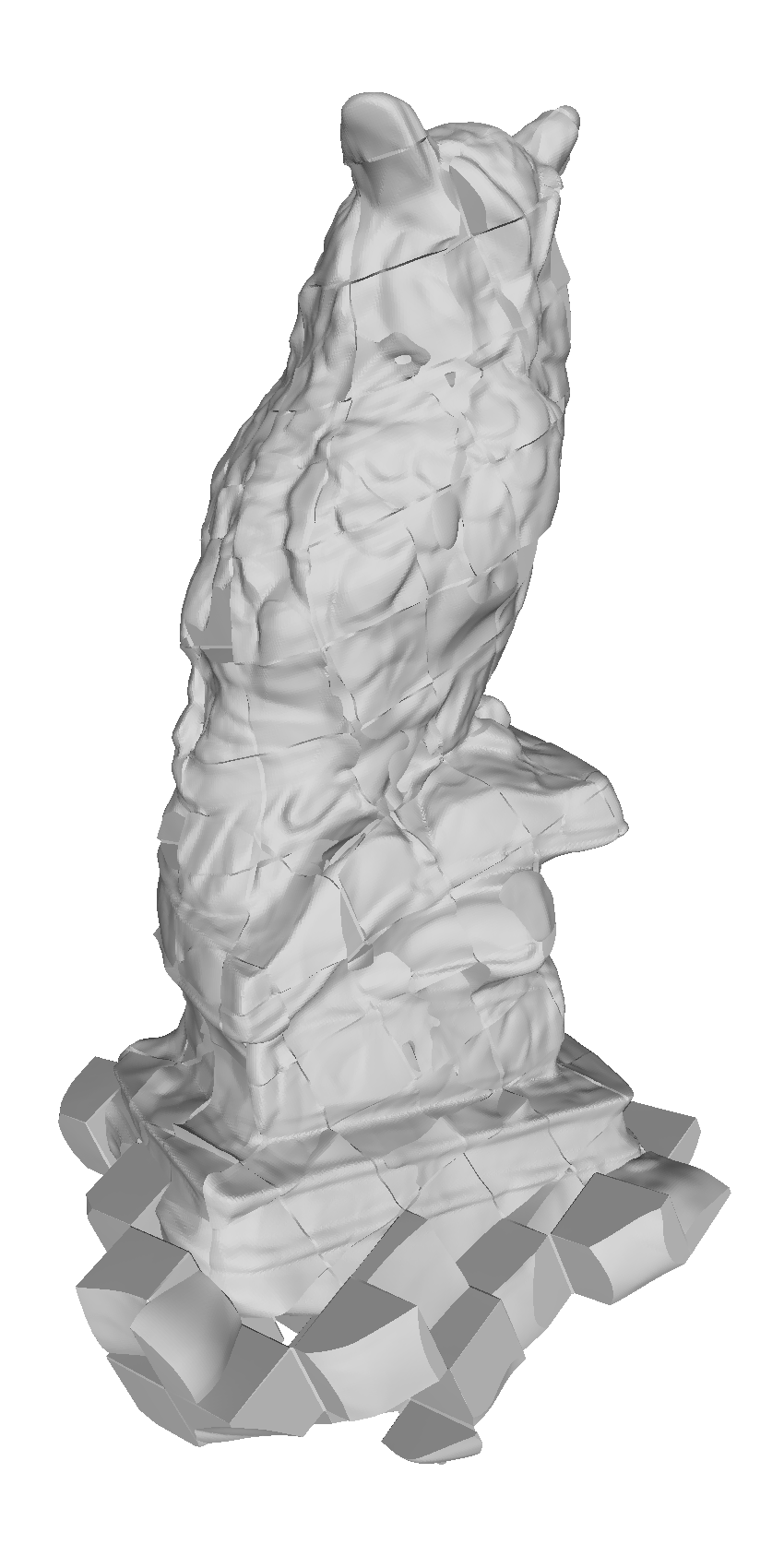}
     \end{subfigure}
     \hfill
     \begin{subfigure}[b]{0.15\textwidth}
         \centering
         \includegraphics[width=\textwidth]{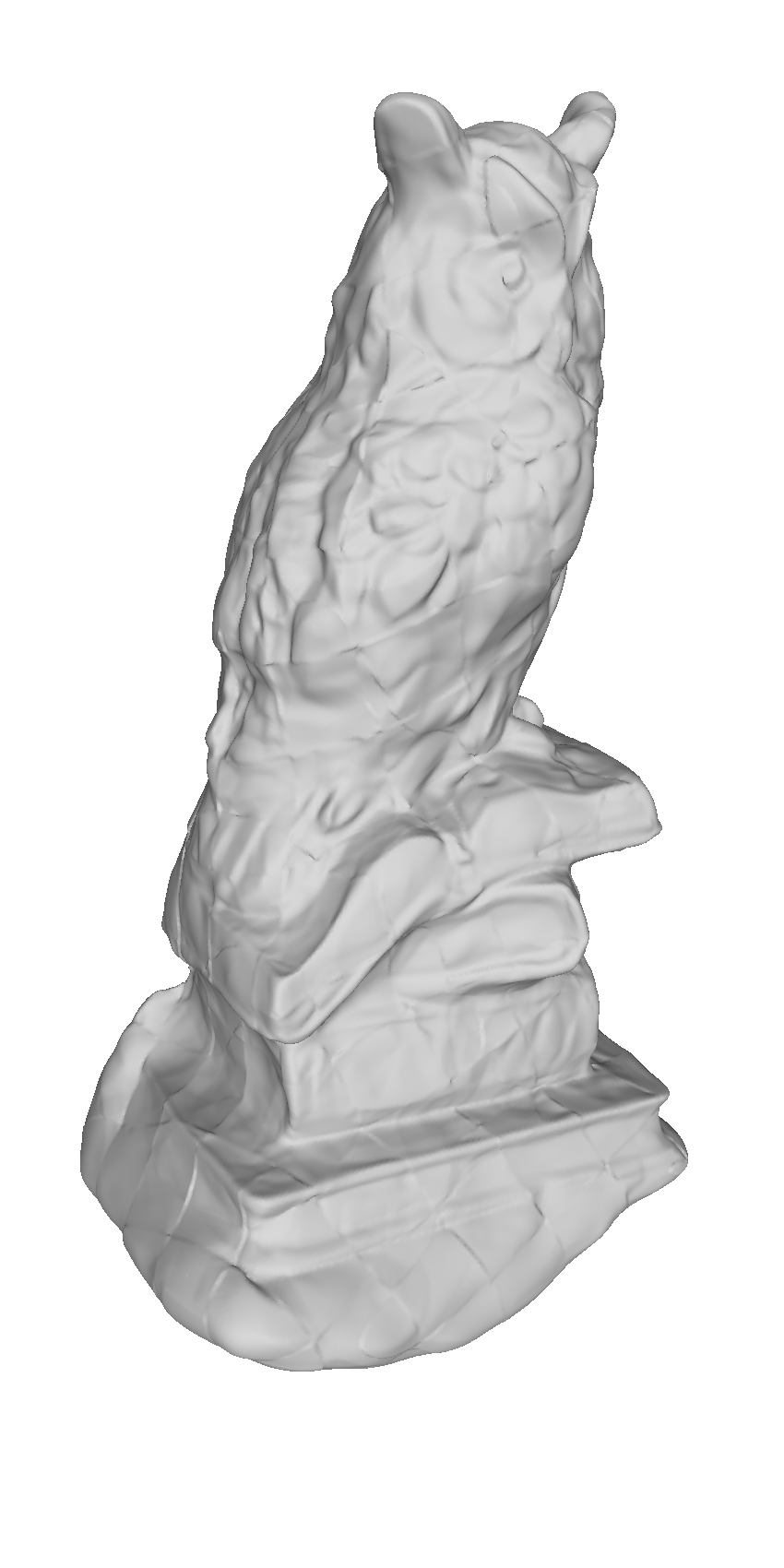}
     \end{subfigure}
     \hfill
     \begin{subfigure}[b]{0.15\textwidth}
         \centering
         \includegraphics[width=\textwidth]{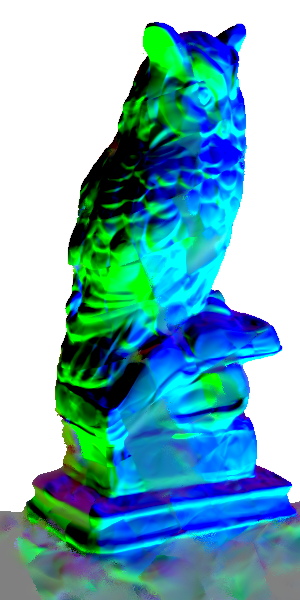}
     \end{subfigure}
     \hfill
     \begin{subfigure}[b]{0.15\textwidth}
         \centering
         \includegraphics[width=\textwidth]{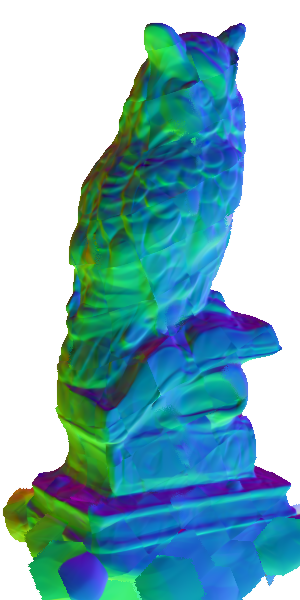}
     \end{subfigure}
     \hfill
     \begin{subfigure}[b]{0.15\textwidth}
         \centering
         \includegraphics[width=\textwidth]{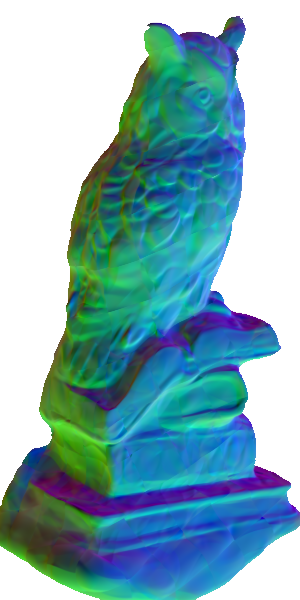}
     \end{subfigure}
     \hfill
    \caption{Surfaces (top row) and normals (bottom row) obtained by fine-tuning (from left to right): without eikonal constraint, with autograd eikonal constraint, and with finite difference eikonal constraint. Note that the output normals are computed as the normalized gradient of the KiloSDF, using autograd in all cases.}
    \label{fig:seams_artifact}
\end{figure}

%-------------------------------------------------------------------------

\section{Experimental results}

In this section, we evaluate KiloNeuS in terms of its representation quality and rendering performance. We aim to quantify our neural representation's ability to learn complex 3D surfaces and their view-independent appearance, by comparing its results for commonly used image-similarity metrics against recent related work. Furthermore, we test the ability of our method to perform surface recovery using ground truth 3D data for our training scenes. 
Lastly, we qualitatively evaluate path-traced renders of KiloNeuS objects in shared scenes containing usual 3D graphics primitives.
We set the Kilo-grid resolution $N$ to 16 for all our experiments, as in Reiser \etal \cite{reiser2021kilonerf}, for a total of $8192$ MLPs (half for SDF and half for color).

We expect our method to perform competitively with the baselines: the Kilo-grid method has already been shown to be capable of improving results quality over the teacher model by Reiser \etal \cite{reiser2021kilonerf}, but we took a further step in improving our method to resolve geometric artifacts introduced by the teacher model (as we explained in Section~\ref{sec:finetune}). However, decoupling surface and color as we do leads to a reduced solution space, given that the surface must be represented by a specific class of functions (\ie, signed distances): therefore, it is possible that our method may not surpass density-field methods (\eg KiloNeRF).

\subsection{Quantitative evaluation}

\paragraph{Surface recovery}  We evaluate the quality of 3D surfaces recovered my multi-view images via KiloNeuS on the DTU \cite{aanaes2016large} dataset, benchmarking its performance against several related methods \cite{wang2021neus, oechsle2021unisurf, mildenhall2020nerf, schoenberger2016mvs}. This experiment mainly serves to show that subdividing space in cubic regions only introduces a controlled amount of artifacts in the recovered surfaces. Table~\ref{tab:surf_rec} shows the complete evaluation.

\begin{table}[h]%
%\begin{minipage}{\columnwidth}
\begin{center}
\caption{Surface recovery evaluation of KiloNeuS and related work. We extract 3D meshes using marching cubes \cite{mcubes} at resolution 512 and run the evaluation procedure described in \cite{aanaes2016large}, which computes a Chamfer distance-based metric between inputs and ground truth 3D point clouds. A lower value indicates higher performance.}
\label{tab:surf_rec}

\begin{tabular}{c || c | c | c | c | c}
    \toprule
    Scene & \textul{Ours} & NeuS & Unisurf & NeRF & Colmap  \\ 
    \midrule
    scan24 & 0.90 & 1.00 & 1.32 & \color{BrickRed}1.90 & \color{OliveGreen}0.81 \\
    scan37 & 1.48 & 1.37 & \color{OliveGreen}1.36 & 1.60 & \color{BrickRed}2.05 \\
    scan40 & 0.87 & 0.93 & 1.72 & \color{BrickRed}1.85 & \color{OliveGreen}0.73 \\
    scan55 & 0.51 & \color{OliveGreen}0.43 & 0.44 & 0.58 & \color{BrickRed}1.22 \\
    scan63 & \color{OliveGreen}0.90 & 1.10 & 1.35 & \color{BrickRed}2.28 & 1.79 \\
    scan65 & 0.95 & \color{OliveGreen}0.65 & 0.79 & 1.27 & \color{BrickRed}1.58 \\
    scan69 & 0.65 & \color{OliveGreen}0.57 & 0.80 & \color{BrickRed}1.47 & 1.02 \\
    scan83 & 1.55 & \color{OliveGreen}1.48 & 1.49 & 1.67 & \color{BrickRed}3.05 \\
    scan97 & 1.11 & \color{OliveGreen}1.09 & 1.37 & \color{BrickRed}2.05 & 1.40 \\
    scan105 & 1.10 & \color{OliveGreen}0.83 & 0.89 & 1.07 & \color{BrickRed}2.05 \\
    scan106 & \color{BrickRed}1.54 & 0.52 & 0.59 & 0.88 & 1.00 \\
    scan110 & 1.87 & \color{OliveGreen}1.20 & 1.47 & \color{BrickRed}2.53 & 1.32 \\
    scan114 & 0.51 & \color{OliveGreen}0.35 & 0.46 & \color{BrickRed}1.06 & 0.49 \\
    scan118 & \color{BrickRed}1.20 & 0.49 & 0.59 & 1.15 & 0.78 \\
    scan122 & 0.66 & \color{OliveGreen}0.54 & 0.62 & \color{BrickRed}0.96 & 1.17 \\
    \midrule
    mean & 1.05 & 0.84 & 1.02 & 1.49 & 1.36 \\
    \bottomrule
\end{tabular}

\end{center}
%\end{minipage}
\end{table}

\paragraph{Volumetric rendering quality}

We conducted our experiments on the DTU \cite{aanaes2016large} dataset, comparing our method against its teacher model NeuS \cite{wang2021neus}, NeRF \cite{mildenhall2020nerf} and KiloNeRF \cite{reiser2021kilonerf}. The results for NeuS were gathered from the pre-trained models published by the authors, which we also used as teacher models for distillation training. Alas, we were not able to reproduce the results of KiloNeRF using their published code, therefore we re-implemented their method in the \texttt{vmap} framework we described in Section~\ref{sec:impl}. We trained NeRF models using the public PyTorch implementation and used them for evaluation and as teachers for KiloNeRF, which we trained with our same settings, applying hierarchical sampling (in order to speed up convergence).
The results of our evaluation confirm our initial guess, improving over both NeuS \cite{wang2021neus} and NeRF \cite{mildenhall2020nerf}. KiloNeRF produces very close results, but performs considerably better over the perceptual metric LPIPS \cite{zhang2018perceptual}. You may refer to Table~\ref{tab:results_table} for a complete overview.

\subsection{Rendering results}

\paragraph{Image quality}

Figure \ref{fig:cornellbox} presents normals, surface color and path-traced renderings of KiloNeuS models. We run our path tracer (in non-interactive mode) with 30 maximum samples per sphere-traced ray and apply denoising on the resulting images for more accurate results. We use Intel Open Image Denoise \cite{oidn} for this purpose.

\paragraph{Performance}

We tested cuNPT on a machine powered by an Intel Xeon E3-1200 v6 processor, 16 GB DDR3 RAM and a NVIDIA GeForce RTX 3090 24 GB, benchmarking its performance. 
To gain a clear indication of the computational cost for the rendering of our neural representation, we limit our performance analysis to scenes whose only element is a KiloNeuS object positioned in front of the virtual camera; doing so, all the rays traced will intersect with its bounding volume going through sphere-tracing steps and, in case of a surface hit, additional MLP inferences. Under this setting, at a resolution of $1280 \times 720$ and 1 sample-per-pixel, the renderer reaches interactive frame rates, averaging $\sim 46$ FPS. However, it is worth noting that the cost might be largely scene- and view-dependent, particularly with multiple shapes and objects being present in the scene. The evaluation of our execution performance can be found in Table~\ref{tab:perf_table}.

\begin{table}[h]%
%\begin{minipage}{\columnwidth}
\begin{center}
\caption{Rendering execution data of our method and related work. We report memory occupancy during rendering (GB) and frames per second (FPS). KiloNeuS maintains a continuous representation while coupling real-time performance and light memory requirements. We report multiple settings for discretization methods (FastNeRF and PlenOctree) in order to show scalability.} 
\label{tab:perf_table}

\begin{tabular}{@{}l || ll@{}}
    \toprule
    \textbf{Method} & \textbf{GB} & \textbf{FPS}  \\ 
    \midrule
    \textul{KiloNeuS} (sphere-traced) & $0.2$ & $46$ \\
    KiloNeRF (MLP grid: $16^3$) & $0.1$ & $45$  \\
    NeuS (sphere-traced) & $0.017$ & $0.1$  \\ 
    NeuS & $0.017$ & $0.01$  \\ 
    NeRF & $0.014$ & $0.01$  \\

    FastNeRF (cache: $1024^3$, factors: 8) & $9.7$ & $238$  \\ 
    FastNeRF (cache: $768^3$, factors: 6) & $4.1$ & $714$  \\ 
    PlenOctree (cache: $512^3$) & $1.9$ & $168$  \\
    PlenOctree (cache: $256^3$) & $0.3$ & $410$  \\
    \bottomrule
\end{tabular}

\end{center}
%\end{minipage}
\end{table}

\subsection{Limitations}

Despite our efforts in this direction, making our surface and color function completely rid of discontinuity artifacts is definitely a non-trivial issue: our model is by definition discontinuous, as it is piece-wise defined on a regular 3D grid. Nevertheless, as we show in Figure~\ref{fig:seams_artifact}, such artifacts are subtle and much less evident than the ones obtained by adopting other straight-forward solutions. It is important to point out that, while other solutions were possible, it was a strict requirement not to over-burden the inference phase, in order to keep rendering real-time: for example, smoothing the output functions would require multiple MLP queries at each point. Our method obtains satisfying results, as attested by our evaluations (see Tables~\ref{tab:surf_rec} and \ref{tab:results_table}), and manages to mitigate this problem completely in the training phase.

By developing our method on top of the NeuS framework, we inherit both its strength and weaknesses: while we are able to reconstruct scenes with high fidelity, foreground/background ambiguities can make our method highly dependent on the completeness of the training views, whereas density field methods can perform reconstruction with fewer or more sparse examples. Furthermore, the volumetric appearance learned by the NeuS Color network, and consequently by our KiloColor network, contains the lighting conditions at the time of dataset image capture baked on top of the true color of the object's surface. This drawback is further addressed in Section~\ref{sec:kiloneus_future_work}, in which we also discuss potential future solutions.

\begin{table}[h]%
%\begin{minipage}{\columnwidth}
\begin{center}
\caption{Rendering quality metrics evaluation of KiloNeuS in contrast to related methods. We evaluate by Peak Signal-to-Noise Ratio (PSNR), Structural Similarity Index (SSIM) \cite{ssim}, and Learned Perceptual Image Patch Similarity (LPIPS) \cite{zhang2018perceptual}. The evaluation dataset is as in Table~\ref{tab:surf_rec}. }
\label{tab:results_table}
\begin{tabular}{l || c c c } 
    \toprule
    \textbf{Method} & PSNR $\uparrow$ & SSIM $\uparrow$ & LPIPS $\downarrow$ \\
    \midrule
    \textul{KiloNeuS} & 30.77 & 0.833 & 0.203 \\
    KiloNeRF  & 30.17 & 0.808 & 0.162 \\
    {NeuS} & 28.55 & 0.820 & 0.220 \\
    {NeRF} & 30.11 & 0.800 & 0.282 \\
    \bottomrule
\end{tabular}

\end{center}
%\end{minipage}
\end{table}

%-------------------------------------------------------------------------

\section{Conclusions}

\subsection{Future work}\label{sec:kiloneus_future_work}

Our work answers to two of the requirements for completely integrating neural representations in classic computer graphics environments, \ie, we propose well-defined 3D objects which can be visualized in real time. We aim to investigate the third requirement in future work, which would be enabling realistic interactions of light with our 3D objects: this point has also been tackled by recent proposals, such as \cite{boss2021nerd, zhang2021nerfactor}, which factorize the appearance of the learned object into additional neural fields encoding \emph{albedo} and  \emph{reflectance} in terms of a spatially varying \emph{Bidirectional Reflectance Distribution Function} (svBRDF). These methods could be used as a template to extend the framework of the NeuS representation, encoding each of these additional properties in different MLPs. From there, they could easily be integrated into our pipeline, allowing the path-tracer to simulate different types of materials. However, increasing the complexity of our representation with additional MLPs appears to be conflicting with the requirement of real-time rendering, as they would translate to several additional MLP queries at inference time. The issue is surely non-trivial and requires additional study, but we believe a success in this direction would be exceedingly valuable, allowing to efficiently perform path-tracing of neural implicit representations.

\subsection{Discussion}
In this paper we proposed KiloNeuS, a new neural object representation specifically designed for its real-time rendering under global illumination. We motivated the rationale behind KiloNeuS design choices and further compared its peculiarities with the characteristics of other well-known representations.  
We then evaluated KiloNeuS, both on a reconstruction quality perspective, showing performance comparable with the state of the art, and on rendering performance through cuNPT, a custom made real-time path-tracer compatible with KiloNeuS objects in scenes containing instances of classic representations. Our performance evaluation of cuNPT showed that our method can achieve competitive efficiency of both memory occupation and rendering time, reaching interactive frame rates.

%-------------------------------------------------------------------------

%%%%%%%%% REFERENCES
{\small
\bibliographystyle{ieee_fullname}
\bibliography{egbib}
}

\end{document}